\documentclass{article}
\usepackage{ijcai19}

\usepackage{times}
\usepackage{soul}
\usepackage{url}
\usepackage[hidelinks]{hyperref}
\usepackage[utf8]{inputenc}
\usepackage[small]{caption}
\usepackage{graphicx}
\graphicspath{{./img/}}
\usepackage{subfigure}
\usepackage{amsmath}
\usepackage{booktabs}
\usepackage{enumerate}
\usepackage{amsfonts}

\title{Less Memory, Faster Speed: Refining Self-Attention Module for Image Reconstruction}

\author{
    Zheng Wang \and
    Jianwu Li \and
    Ge Song \and
    Tieling Li
    \affiliations
    Beijing Institute of Technology \emails
    \{2120171117, ljw, 2120171117, 2220170569\}@bit.edu.cn
}

\begin{document}

\maketitle

\begin{abstract}
Self-attention (SA) mechanisms can capture effectively global dependencies in deep neural networks, and have been applied to natural language processing and image processing successfully. However, SA modules for image reconstruction have high time and space complexity, which restrict their applications to higher-resolution images. In this paper, we refine the SA module in self-attention generative adversarial networks (SAGAN) via adapting a non-local operation, revising the connectivity among the units in SA module and re-implementing its computational pattern, such that its time and space complexity is reduced from $\text{O}(n^2)$ to $\text{O}(n)$, but it is still equivalent to the original SA module. Further, we explore the principles behind the module and discover that our module is a special kind of channel attention mechanisms. Experimental results based on two benchmark datasets of image reconstruction, verify that under the same computational environment, two models can achieve comparable effectiveness for image reconstruction, but the proposed one runs faster and takes up less memory space.
\end{abstract}

\section{Introduction}
It has been proven that capturing long-range dependencies in deep neural networks is helpful in improving their application effects, especially for image processing. Previously, long-range dependencies of images are captured by large receptive fields formed by deep stacks of convolutional operations \cite{n14,n30}. However, since convolutional operations only focus on a local neighborhood, long-range dependencies need to be captured via applying them repeatedly. There are two main limitations of repeating such local operations. First, it causes difficulty in optimization \cite{n23,n21}. Second, it is computationally inefficient.

Recently, Wang et al.~\cite{nonlocal} applied non-local operations to capture efficiently long-range dependencies by using non-local mean operations~\cite{n4} in deep neural networks. The non-local operations compute the response at a position as a weighted sum of the features at all positions in input feature maps. 
Zhang et al.~\cite{sagan} proposed Self-Attention Generative Adversarial Networks (SAGAN) which use one of non-local operations to implement self-attention modules.
SAGAN obtains state-of-the-art results on ImageNet dataset \cite{imagenet}. SAGAN is the first to combine Generative Adversarial Networks with Self-Attention (SA) mechanism, and generates a new solution of computer vision, especially for image reconstruction. However, the SA module of SAGAN has two limitations:
\begin{enumerate}[1.]
\item The SA module is hard to be employed on bigger datasets with higher dimensions, since it has a space complexity of $\text{O}(n^2)$. It limits many applications of self-attention in computer vision. For example, based on \cite{large}, the performance of generative tasks has positive correlation with the batch size of its training and a high space complexity will restrict the increase of batch size. 
\item It also has a time complexity of $\text{O}(n^2)$. Although it improves the quality of image generation, using the self-attention module brings huge time costs in both testing and training phases.
\end{enumerate}

In this paper, we propose a new self-attention module for overcoming these limitations. Compared with the original module, in theory, the new self-attention module has the space and time complexity of $\text{O}(n)$ instead of $\text{O}(n^2)$, and in practice, the time and memory can be saved up $30\% \sim 50\%$ (depending on the dimensions of input data and the structure of networks), while obtaining comparable performance with the vanilla SA. Further, we can introduce the proposed self-attention mechanism into GANs or other deep neural networks to reduce their computational costs. Our contributions include:
\begin{enumerate}[1.]
\item We implement a new self-attention module which has a time and space complexity of $\text{O}(n)$.
\item We analyze the proposed module from the view point of channel attention \cite{ca} and further compare them.
\item We provide two experiments to verify the performance of the proposed module for image reconstruction.
\end{enumerate}

%This paper is organized as following. Section 2 discusses some related works. Section 3 mainly shows the theory of our proposed module including the its mathematical formula derivation and the principle behind itself. Section 4 discusses the time and space complexity of the two self-attention modules and the advantages of our module. Section 5 includes a completion experiment and a generative experiment. The completion experiment can be seen as an ablation study to show our module has a great potential in where a problem need process long-range dependency. The generative experiment is to verify the effectiveness of the two modules while to compare the time spent and memory consumed under the same condition. 

\section{Related Works}
\subsection{Self-attention (SA)}
The advantages of self-attention in capturing global dependencies make attention mechanisms become an integral part of modules \cite{sa2,sa34,sa36,sa6}. In particular, self-attention \cite{sa4,sa20} computes the response at a position as a weighted sum of the features at all positions in input feature maps. By adding self-attention module into an autoregressive module for image generation, Parmar et al. \cite{sa21} propose an image transformer module. Wang et al. \cite{nonlocal} formalize self-attention as non-local operations inspired by non-local mean filter \cite{n4}. Based on non-local operations, Zhang et al. \cite{sagan} present Self-Attention Generative Adversarial Network (SAGAN) to generate images based on ImageNet \cite{imagenet} and obtain 27.62 of Fr\'echet inception score (FID) compared with previous state-of-the-art 18.65.

\subsection{Self-Attention Module in SAGAN}
Self-attention module in SAGAN as shown in Figure \ref{strua}, is based on non-local neural networks inspired by non-local mean filters. Following the baseline, the non-local operation can be defined generally as:
\begin{equation}
 out_i=\sum_{\forall j} \frac {1} {\mathcal{C}(x)}  f(z_i,y_j)g(x_j), \label{base}
\end{equation}
where $x \in \mathbb{R}^{N\times C}$ is the matrix of image features and $C$, $N$ denote the numbers of channels and elements of one channel, respectively. $z=W_zx$, $y=W_yx$ are two embeddings of $x$, and $W_z$ and $W_y$ are their embedding matrices. $out$ is output signal. The pairwise function $f$ computes a relationship between the $i$th and the $j$th elements of $x$. The unary function $g(x)$ represents features of $x$ and the equation is normalized by a factor $\mathcal{C}(x)$.
Based on the non-local embedded operation, self-attention mechanism is only a special case. The implementation is shown as:
\begin{equation}
out=softmax(zy^T)\phi(x), \label{sagan}
\end{equation}
where $\phi(x)=W_\phi x$ is a linear feature transform of $x$. In this case, $softmax$ function can be seen as $\frac 1 {\mathcal{C}(x)}$, and $f$ becomes the cosine similarity.

\section{Improved Self-Attention Module}
Using SA modules, SAGAN reduces the FID from 27.62 to 18.65 on the challenging ImageNet dataset. The results show the SA mechanism has a great potential in image reconstruction. However, the time and space complexity of the SA module of SAGAN are $\text{O}(n^2)$ (in Section 4). Especially in space complexity, if we enlarge the resolution of generative images from 128 * 128 to 256 * 256, the consumed memory will be enlarged from $128^4$ to $256^4$. Huge consumption of computing resources affects the applications of SA seriously, even with the help of GPU computers.

\subsection{The Novel Self-attention Module}
\begin{figure*}[ht]
\centering
\subfigure[The self-attention module of SAGAN ]{\label{strua}\includegraphics[width=0.77\textwidth]{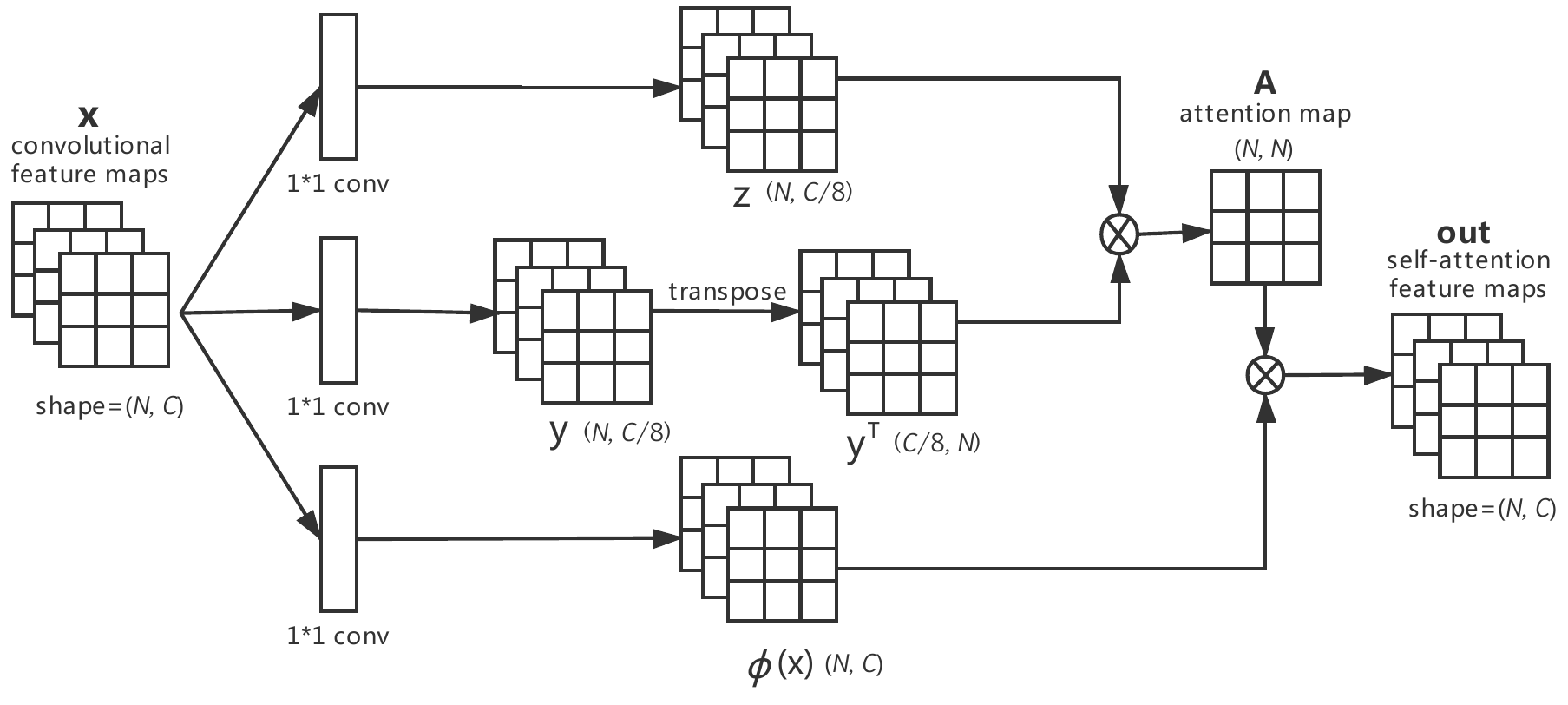}}
\subfigure[The proposed self-attention module]
{\label{strub}\includegraphics[width=0.77\textwidth]{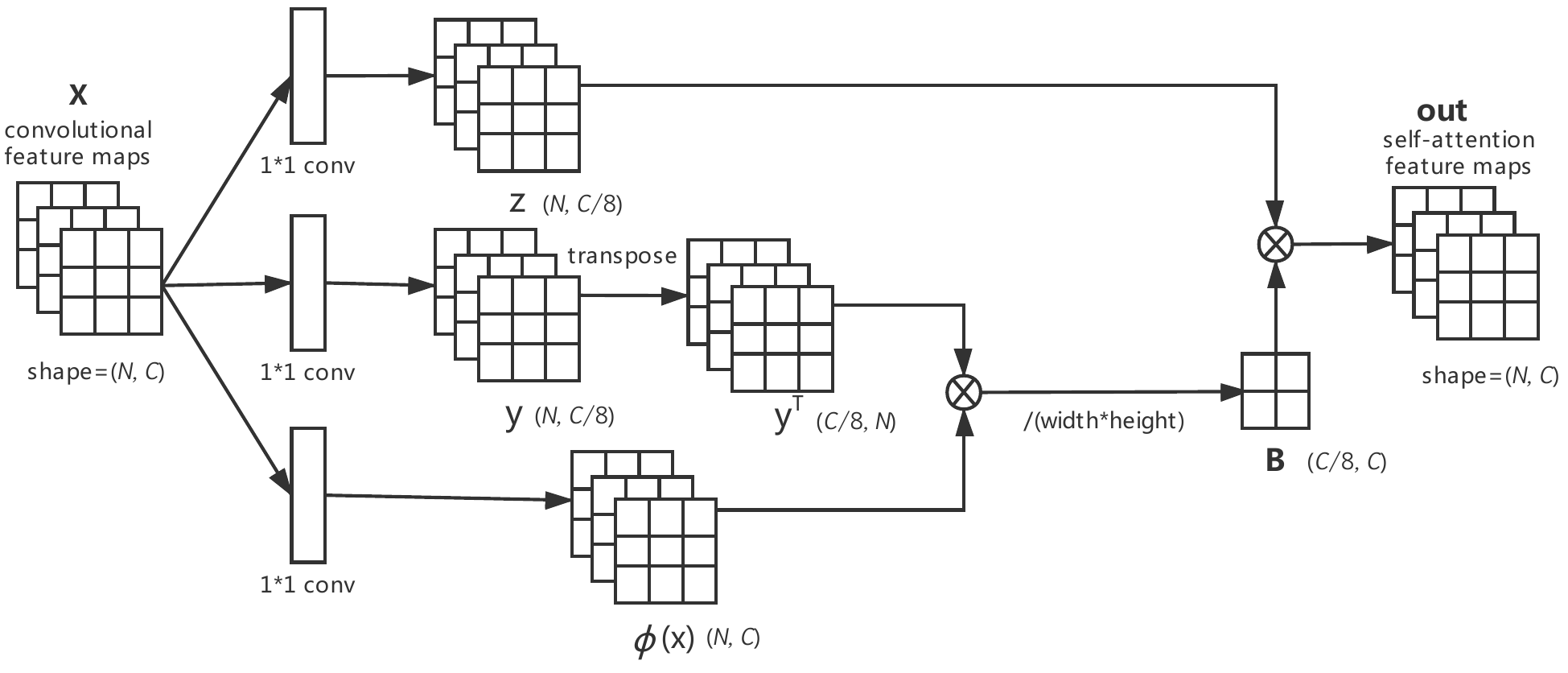}}

\caption{The structures of two kinds of SA modules where $\otimes$ represents matrix multiplication and $/$ is scalar division.}
\label{stu} 
\end{figure*}

From Figure \ref{strua}, the reason why the SA module of SAGAN consumes so much memory and time lies in the computation of attention map $A$. The attention map calculates any pair of elements in its input. Hence, the key to reduce the computational complexity is to modify the computational way of attention maps. Inspired by the associativity of matrix multiplication, we find if we first calculate $y^T\phi(x)$ in Equation (\ref{sagan}), we will obtain a $(C/8, C)$ matrix rather than a $(N, N)$ matrix. In convolutional operations, $C$ is a hyper-parameter and generally $N \gg C$. Obviously, the revised computation can reduce computational complexity to a large extent. However, $softmax$ is not a linear function, we need to use a linear function to replace $softmax$ function.

Following the statement of \cite{nonlocal}, although recent self-attention modules mostly take $softmax$ as the normalization factor, \cite{nonlocal} uses two alternative versions of non-local operations to prove the nonlinear attentional behavior is not essential. Further, they also make experiments to verify that the results of those versions are comparable in video classification and image recognition. Thus we can employ one of the versions to rewrite the self-attention module: 
\begin{equation}
out=\frac {zy^T\cdot \phi (x)} {N}.\label{ostart}
\end{equation}
We continue to rewrite Equation (\ref{ostart}) by the associativity of matrix multiplication: 
\begin{equation}
out=\frac {zy^T\cdot \phi (x)} {N}=z\cdot\frac {(y^T\phi (x))} {N},\label{ofinal}
\end{equation}
where $\frac {(y^T\phi (x))} {N}$ is of a space complexity of $  \text{O}(1)$. Following Equation (\ref{ofinal}), we design a network, the structure of which is shown in Figure~\ref{strub}. The structures in Figure \ref{strub} and Figure \ref{strua} are different, and also have different meanings, which will be analyzed in the following section.

\subsection{The Principle Behind the Module}
In Figure \ref{stu}, we use a schematic diagram to compare the two modules of self-attention. Generally, $N\gg C$. For simplification, we set $N=4,C=2$, and the number of channels keeps invariant in transforms, i.e., the number of channels is not divided by 8. The explanation starts from calculating an element of matrix $out$ in different ways.
\begin{equation}
% \sum_{k=1}^{N=4}A_{ik}{\lbrack\phi(x)\rbrack}_{kj} \\
\begin{aligned}
out_{ij}&=A_{i}{\lbrack\phi(x)\rbrack}_{j}'&= \sum_{k=1}^{N=4}A_{ik}{\lbrack\phi(x)\rbrack}_{kj},\label{ea}
\end{aligned}
\end{equation}

\begin{equation}
\begin{aligned}
out_{ij}&=z_{i}B_{j}'\\&=z_{i1} \frac  { y^T_1{\lbrack\phi(x)\rbrack}_{j}'} {N} + z_{i2} \frac  { y^T_2{\lbrack\phi(x)\rbrack}_{j}'} {N} \\ &=z_{i1}t_{1j}+z_{i2}t_{2j}=\sum_{k=1}^{C=2}t_{kj}z_{ik},\label{eb}
\end{aligned}
\end{equation}
where $t_{ij}=\frac  { y^T_i{\lbrack\phi(x)\rbrack}_{j}'} N$ and $A_{ij}=\frac  { exp(z_iy_j)} {\sum_{\forall(i,j)}{exp(z_iy_j)}}$. $A_i$, $z_i$ represent row vectors consisting of the elements in the $i$th row of $A$, $z$, respectively. And $B_i'$, ${\lbrack\phi(x)\rbrack}_{j}'$ denote column vectors consisting of the elements in the $i$th column of $B$, $z$, respectively. %$out_{ij}$, ${\lbrack\phi(x)\rbrack}_{ij}$, $y_{ij}$, $z_{ij}$, $out_{ij}$ represent the elements of itself located in the $i$th row, the $j$th column.

For the original SA module, calculating an element of $out$ is to compute a cosine similarities between the element and all elements of input. Through Equation (\ref{ea}), $out_{ij}=A_i[\phi(x)]_j$ where $A_{ij}\approx \frac  { z_iy_j} {\|z_i\|\|y_j\|}$ represents the cosine similarity between the $i$th element and the $j$th element, $out_{ij}$ is computed by a weighted sum of all elements and the weights depend on cosine similarity which is defined as a product of two normalized vectors. Matrix $A$ is the weight matrix whose element $A_{ij}$ is the cosine similarity between $z_i$ and $y_j$.

However, the attention map $B$ is produced by inner product, which means that $B_{ij}$ cannot represent the cosine similarity between $z_i$ and $y_j$. Like the analysis of attention map $A$, we also calculate an element of $out$ in the proposed module. Through Equation (\ref{eb}), the results of the proposed module are formed by a weighted sum of all channels, and the weights depend on a similarity $t$ corresponding to cosine similarity. That means, the proposed module computes the similarity of every two channels of all elements (e.g. $out_1'$) rather than every two elements of all channels (e.g. $out_1$). The principle behind the proposed module is similar to that of channel attention modules \cite{ca,casr} that make important channels be focused. To explain it more clearly, we first make transformation of Equation (\ref{eb}):
\begin{equation}
out_i'=z_1' t_{1i}+z_2' t_{2i},
\end{equation}
where $out_i'$, $z_i'$ represent column vectors consisting of the elements in the $i$th channel, Since $z_i'$ is produced by $1*1$ convolutional operation, we can obtain:
\begin{equation}
z_i'=\frac {w_i} {w_j} \cdot z_j',
\end{equation}
where $w_i$, $w_j$ denote weights of the $i, j$th $1*1$ convolutional operations, respectively. Further making $z_i'=w_a\cdot z_1'=w_b\cdot z_2'$ where $w_a= \frac {w_1} {w_i},  w_b=\frac {w_2} {w_i}$, we obtain:
\begin{equation}
\begin{aligned}
out_i'&=z_1't_{1i}+z_2't_{2i}\\ &=w_at_{1i}\cdot z_i'+w_b t_{2i}\cdot z_i'\\ &=(w_a t_{1i}+w_b t_{2i})\cdot z_i'=cz_i'~~\text{where} ~c \in \mathbb{R}.\label{efinal}
\end{aligned}
\end{equation}
Equation (\ref{efinal}) shows that the proposed module aims to assign a weight for every channel to make some important channels be focused. The weight $c$ is composed of two parts. The first part includes $t_{1i}$ and $t_{2i}$ calculated by global information and the second part includes $w_a$ and $w_b$ which are learnable parameters.

\subsection{Comparation with Channel Attention}
\begin{figure}[h]
\centering
\subfigure[channel attention]{\centering\label{ca1}\includegraphics[width=0.21\textwidth]{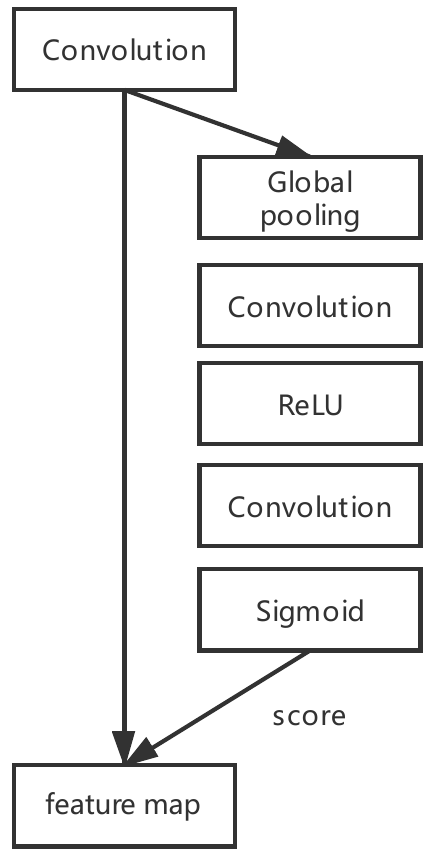}}
\subfigure[channel attention (ours)]{\centering\label{ca2}\includegraphics[width=0.26\textwidth]{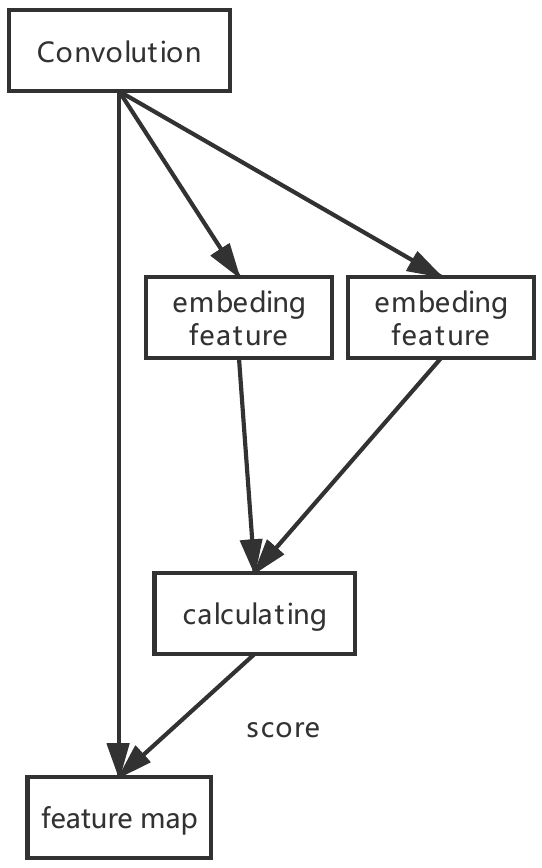}}
\caption{Two channel attention modules.}
\label{ca}
\end{figure}
Channel attention (CA) is firstly proposed in \cite{ca}, which generates different scores for each channel-wise feature. As shown in Figure \ref{ca1}, the original channel attention uses global average pooling to obtain global information, which is defined as:
\begin{equation}
m_c=\frac 1 N \sum_{i, j} x_{i,j}
\end{equation}
where $m_c \in \mathbb{R}^{1\times1\times C}$ represents global information captured by the CA module. Then the information is processed by two non-linear transformations to obtain the score $s_c$ for each channel-wise:
\begin{equation}
s_c=\sigma(w_2\delta(w_1 m_c))
\end{equation}
Finally, the initial feature map $x$ is multiplied by $s_c$:
\begin{equation}
out = s_c*x
\end{equation}
The CA module in \cite{ca} and the proposed module have similar purposes which assign a learnable weight to each channel. There are two differences between the two CA modules. The first difference lies in the way to obtain global information. The original CA module uses a global pooling to add all elements of a channel, while our module computes the relationship between any pair of channels. Secondly, the original module is a nonlinear module, whereas our module is linear. The nonlinearity of the original module is from its nonlinear active function. However, the linearity of  our module does not affect its effect. For instance, a network generally does not use only one SA module like Figure \ref{struc} and throughout those modules, there are some nonlinear layers to do the nonlinear transformation.

\section{Complexity of Computation}
As shown in Figure \ref{strua}, the original self-attention module which computes an attention map that explicitly represents the relationships between any two positions in the convolutional feature map $x$, so there are $N*N$ elements of attention map $A$. Since there is a quadratic relationship between the elements of image feature $x$ and the size of attention map $A=zy^T$, the space complexity is $\text{O}(n^2)$. 
%At the same time, if considering the time complexity is $O(m*n*p)$ for a multiplication between a $(m*n)$ matrix and a $(n*p)$ matrix, 
The time complexity of the original module is also $\text{O}(n^2)$, since it computes multiplications between the attention map and convolutional feature map, whose dimensions are $(N*N)$ and $(N*C)$, respectively.

Our self-attention module has the space and time complexity of $\text{O}(n)$. According to Figure \ref{strub}, there are two multiplications. The first multiplication between a $(C/8*N)$ matrix $y^T$ and a $(N*C)$ matrix $\phi(x) $ produces a $(C/8*C)$ matrix $B$ and it times a $(C/8*N)$ matrix $z$. Hence, the time and memory space costs are linear relationships with $N$, and correspondingly, both the time and space complexity are $\text{O}(n)$. 

Via reducing memory space and computational time, we can apply the self-attention mechanism to the fields of image reconstruction including image completion, super resolution, etc., which need relatively more computational resources.

\section{Experiments}
We provide two experiments to measure and compare the proposed module with the original one. The first experiment is an ablation study which applies our module to complete some images with large margin missing, in order to measure the effectiveness for capturing long-range dependencies. The second experiment is about image generation, in order to prove that the refined module can obtain comparable results with the original module in SAGAN but consume less memory space and running time. The two experiments are carried out on a platform of NVIDIA GTX 1080ti GPU, 32 GB RAM and i7-7700k CPU.
\subsection{Ablation Study}
The experiment is to verify the ability of our self-attention module to capture long-range dependencies for image reconstruction. The purpose of the experiment is to complete an image (e.g. Figure~\ref{cr}) which is cut 1/4 both on its left and right side (e.g. Figure~\ref{cri}). We will observe the influences after replacing two convolutial layers of a standard GAN by our SA module (illustrated in Figure \ref{struc}).
\subsubsection{Details of Implementation}

\begin{figure*}[h]
\centering
\includegraphics[width=0.75\textwidth]{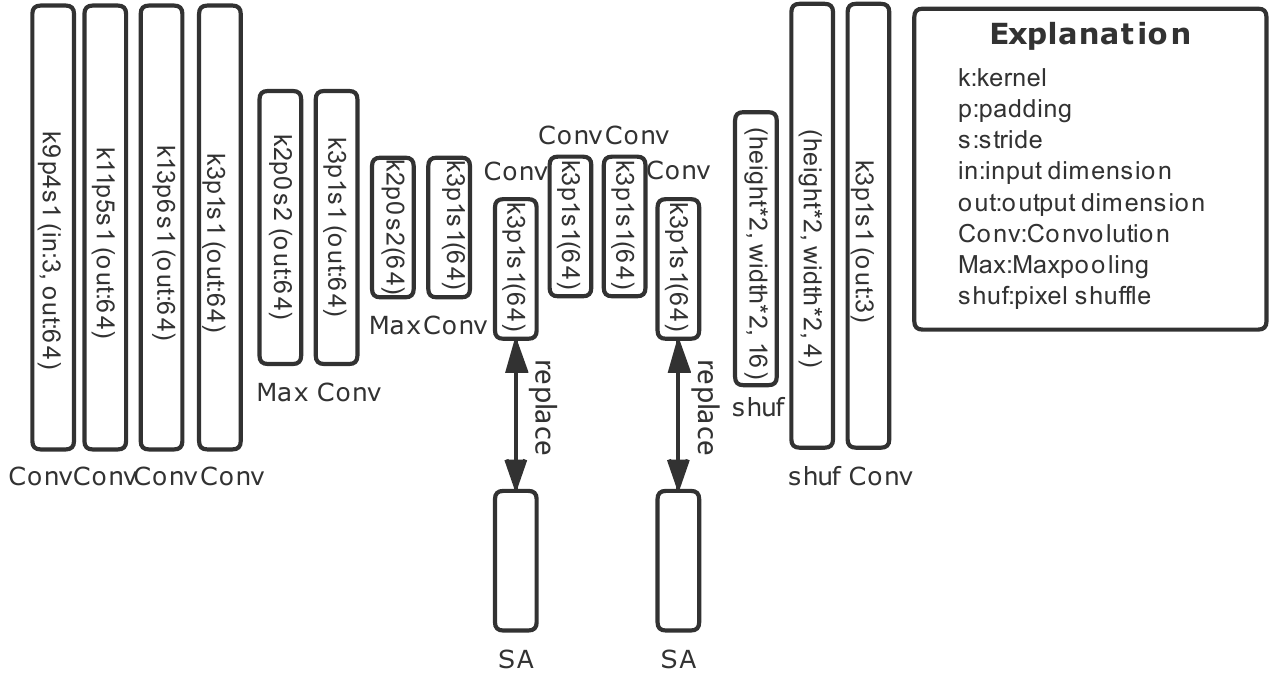}
\caption{The structure of generator in the experiments.}
\label{struc}
\end{figure*}

The used network is a basical super resolution generative adversarial network (SRGAN) \cite{srgan}. All settings of the experiment are inherited from \cite{srgan} except that we re-implement the structure of generators like Figure \ref{struc} and add spectral normalization \cite{sn} for every convolutional layer of the generator to stabilize the training phase. Additionally, the dataset used is a simple coast dataset \footnote{It can be downloaded on \url{http://cvcl.mit.edu/scenedatabase/coast.zip}}, the images of which are resized into 256*256 in pre-processing.

\subsubsection{Experimental Results}
\begin{figure*}[!h]
\centering
\subfigure[Ground truth]{ \label{crg}
\begin{minipage}[b]{0.23\textwidth}
\centering
\includegraphics[width=1\textwidth]{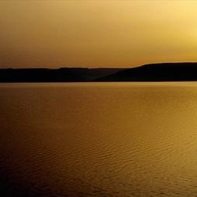} \\
\includegraphics[width=1\textwidth]{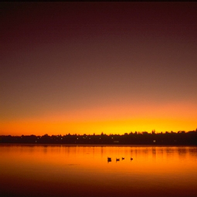}
\end{minipage}
}
\subfigure[Input]{ \label{cri}
\centering
\begin{minipage}[b]{0.23\textwidth}
\includegraphics[width=1\textwidth]{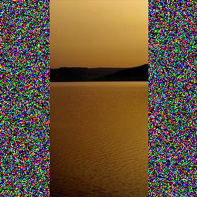} \\
\includegraphics[width=1\textwidth]{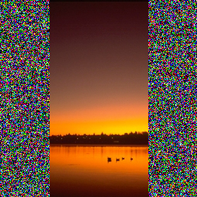}
\end{minipage}
}
\subfigure[The results of standard GAN]{ \label{crgan}
\centering
\begin{minipage}[b]{0.23\textwidth}
\includegraphics[width=1\textwidth]{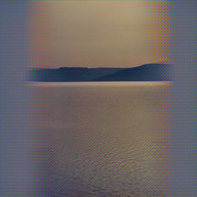} \\
\includegraphics[width=1\textwidth]{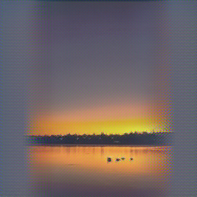}
\end{minipage}
}
\subfigure[The Results of the Proposed Module]{ \label{cro}
\centering
\begin{minipage}[b]{0.23\textwidth}
\includegraphics[width=1\textwidth]{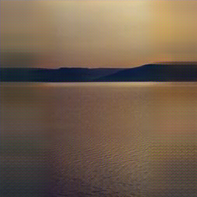} \\
\includegraphics[width=1\textwidth]{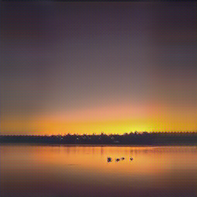}
\end{minipage}
}
\caption{The results of image completion. We train GANs to complete images in (b). (c) shows results of a standard GAN. When two convolutional layers are replaced by our SA model, we obtain the results in (d).} \label{cr}
\end{figure*}

The image reconstruction from an image with large margins missing cannot achieve satisfactory results if a network is hard to handle long-range dependencies as shown in Figure \ref{crgan}. This experiment is to inspect whether our SA model still has an ability to capture long-range information after structural transforms. Figure \ref{cro} shows the results of our SA model. Without our SA module, the regions to be completed are hard to receive valid signals provided by residual regions (Figure \ref{crgan}). Hence the experiments verify that our SA module inherits the ability from the original self-attention module.

\subsection{Generative Experiments}
Furthermore, we compare the performance of two modules in a real computational environment. To ensure a fair comparison, we choose three networks only with single difference in implementation of SA model. The three networks are self-attention generative adversarial network (SAGAN), improved self-attention generative adversarial network (ISAGAN) and standard generative adversarial network (SGAN). SAGAN uses its original implementation in \cite{sagan}\footnote{The code can be downloaded on \url{https://github.com/heykeetae/Self-Attention-GAN}} and ISAGAN uses the structure of SAGAN but replaces the SA models by the proposed one. SGAN replaces SA modules by convolution layers as a criterion. We train the three networks based on two benchmark datasets $CIFAR-10$ and $CelebA$ and evaluate them by Fr\'echet inception score \cite{fid} (FID) (the lower is the better).
Generally, a generative task is evaluated by Inception score \cite{is} and Fr\'echet inception score (FID), but based on \cite{isn},  Inception score is misleading when a generative network is not trained on ImageNet.
\subsubsection{Details of Implementation}
All the generative models are designed to generate 64*64 images. By default, the batch size is 64 and other hyper-parameters of discriminators, generators and optimizers are inherited from SAGAN.
\subsubsection{Experimental Results}
\begin{table}[htbp]
\caption{The results based on two datasets}
\centering
\begin{tabular}{lccc}
\hline 
FID&	SAGAN&	ISAGAN&	SGAN \\
\hline 
$CelebA$&8.864	&8.723	&6.394\\
\hline  
$CIFAR-10$&	12.739	&12.236	&16.066\\
\hline 
\end{tabular}
\label{ger1}
\end{table}
The results of the three networks (SAGAN, ISAGAN, GAN) are tabulated in Table \ref{ger1}. Compared with SGAN, SAGAN and ISAGAN achieve comparable effects. Concretely, trained on $CelebA$, a human face dataset, both SAGAN and ISAGAN degrade the generative quality, since generating human face may depend more on local features than global features and thus the advantages of the self-attention mechanism are not helpful and even have some interferences. Whereas, trained on $CIFAR-10$, a multiple classes dataset, the two self-attention modules improve the quality with the advantage of their long-range dependencies. Through the generative experiments and the formula derivation, we can infer the two self-attention modules are comparable in effectiveness. 

Furthermore, we need to evaluate the costs of time and space of our proposed module. Table \ref{ger2} shows the time spent in forward (every 30k images) and backward (every 10 * batch size images) propagation.
\begin{table}[htbp]
\caption{The time spent in forward and backward propagation on $CIFAR-10$.}
\centering
\begin{tabular}{lccc}
\hline 
time (in seconds)&	SAGAN&	ISAGAN&	SGAN \\
\hline 
forward&	1.077	&0.762&	0.564\\
\hline  
backward&	5.992&	3.049&	2.390\\
\hline 
\end{tabular}
\label{ger2}
\end{table}
The reason why the speeds of ISAGAN on two propagations are faster than those of SAGAN is mainly that our self-attention module avoids the large-scale matrix multiplication. In forward propagation, the large-scale matrix multiplication happens to more units, and in backward propagation, it needs to do differentiation on larger computational graph.
About memory space usage, Table \ref{ger3} shows the training of SAGAN and ISAGAN with different batch sizes, respectively, where `no' means that a module cannot be run in our environment and `ok' represents the opposite.
\begin{table}[htbp]
\caption{Training of the three models with different batch sizes on $CIFAR-10$}
\centering
\begin{tabular}{lccc}
\hline 
Batch size&	SAGAN&	ISAGAN&	SGAN \\
\hline 
256&	ok&	ok&	ok\\
\hline  
512&	no&	ok&	ok\\
\hline 
1024&	no&	ok&	ok \\
\hline 
\end{tabular}
\label{ger3}
\end{table}
Since increasing the batch size needs more memory space for training, we can also infer that ISAGAN uses less memory space than SAGAN.
\section{Conclusion }
We improve the original self-attention module to reduce time and space complexity. Due to less memory space consumption, our self-attention module can be used in image reconstruction which often needs to process higher dimensions data. Theoretically, our SA module is a special kind of channel attention mechanisms. Experimental results verify that using our self-attention module can obtain comparable effects with the vanilla one but use less time and memory space. In future work, we will apply the proposed module of self-attention to other deep learning tasks, beyond image reconstruction.

\bibliographystyle{named}
\bibliography{ijcai19}
\end{document}